\documentclass[11pt, leqno]{article}
\usepackage[english]{babel}

\usepackage[utf8]{inputenc}
\usepackage[T1]{fontenc}
\usepackage{dsfont}
\usepackage{hyperref}       
\usepackage{url}            
\usepackage{booktabs}       
\usepackage{amsfonts}       
\usepackage{nicefrac}       
\usepackage{amsmath, amssymb}
\usepackage[a4paper,hscale=0.72,vscale=0.8,centering]{geometry}
\usepackage{setspace}
\usepackage{authblk}
\usepackage{amsfonts}
\usepackage{bbm}
\usepackage{dsfont}
\usepackage{graphicx}
\usepackage{float}
\usepackage{xcolor}
\usepackage{wrapfig}
\usepackage{tikz}
\usepackage{pst-node}
\usepackage{listings}
\usepackage{fancyvrb}
\usetikzlibrary{fit,positioning,bayesnet}
\usetikzlibrary{automata,arrows}
\usetikzlibrary{backgrounds}

\usepackage{subfigure}
\usepackage{amsfonts}
\usepackage{footmisc}
\usepackage{bm}
\usepackage{diagbox}
\usepackage[linesnumbered,ruled,vlined]{algorithm2e}
\usepackage[numbers,sort&compress]{natbib}
\usepackage{caption}
\usepackage{natbib}
\usepackage{hyperref}
\definecolor{bred}{rgb}{0.8,0,0}
\hypersetup{colorlinks,linkcolor={blue},citecolor={black},urlcolor={blue}}
\setcitestyle{square,numbers,comma}
\usepackage{cleveref}

\def \0{\mathbf 0}

\def\b0{\mathbf 0}

\usepackage{xcolor}
\usepackage{bbm}

\usepackage{tikz}
\usetikzlibrary{shapes, positioning, calc}

\usepackage{parskip}
\usepackage{titlesec}

\titlespacing\section{0pt}{16pt plus 4pt minus 2pt}{6pt plus 2pt minus 2pt}
\titlespacing\subsection{0pt}{16pt plus 4pt minus 2pt}{6pt plus 2pt minus 2pt}
\titlespacing\subsubsection{0pt}{16pt plus 4pt minus 2pt}{6pt plus 2pt minus 2pt}

\titleformat{\section}[hang]{\normalfont\Large\bfseries}{\thesection}{0.5em}{}[]
\titleformat{\subsection}[hang]{\normalfont\Large\bfseries}{\thesubsection}{0.5em}{}[]
\titleformat{\subsubsection}[hang]{\normalfont\large\bfseries}{\thesubsubsection}{0.5em}{}[]

\ifpdf
\hypersetup{
  pdftitle={A Primer on Variational Inference for Physics-Informed Deep Generative Modelling},
  pdfauthor={A. Glyn-Davies, A. Vadeboncoeur, OD. Akyildiz, I. Kazlauskaite, M. Girolami }
}
\fi

\usepackage[acronym]{glossaries}
\newacronym{wrm}{WRM}{weighted residual method}
\newacronym{fe}{FE}{finite element}
\newacronym{uq}{UQ}{uncertainty quantification}
\newacronym{pinn}{PINN}{physics-informed neural networks}
\newacronym{ml}{ML}{machine learning}
\newacronym{phi-ml}{Phi-ML}{physics-informed machine learning}
\newacronym{vi}{VI}{variational inference}
\newacronym{bip}{BIP}{Bayesian inverse problem}
\newacronym{pde}{PDE}{partial differential equation}
\newacronym{ode}{ODE}{ordinary differential equation}
\newacronym{sde}{ODE}{stochastic differential equation}
\newacronym{elbo}{ELBO}{evidence lower bound}
\newacronym{vae}{VAE}{variaitonal auto-encoder}
\newacronym{kl}{$\mathrm{KL}$}{Kullback–Leibler}
\newacronym{mcmc}{MCMC}{Markov chain Monte Carlo}
\newacronym{em}{EM}{Expectation Maximisation}
\newacronym{dgp}{DGP}{deep generative prior}

\usepackage{mathtools}

\title{{\scshape{Probabilistic Super-Resolution for High-Fidelity Physical System Simulations with Uncertainty Quantification}}}

\newcommand{\cblue}{\textcolor{blue}}

\author[$\cblue{\dagger} $]{Pengyu Zhang \footnote{Corresponding author. Email address: \href{mailto:pz281@cam.ac.uk}{pz281@cam.ac.uk} }}
\author[$\cblue{\dagger}$]{Connor Duffin}
\author[$\cblue{\dagger}$]{Alex Glyn-Davies}
\author[$\cblue{\dagger}$]{\\ Arnaud Vadeboncoeur}
\author[$\cblue{\dagger} \S$]{Mark Girolami}

\affil[$\cblue{\dagger}$]{Department of Engineering, University of Cambridge}
\affil[$\cblue{\S}$]{The Alan Turing Institute}

\date{}
\date{\small \textbf{Keywords}: Super Resolution; Uncertainty Quantification; Bayesian Methods}
\begin{document}
\setstretch{1.}

\maketitle

\begin{abstract}
Super-resolution (SR) is a promising tool for generating high-fidelity simulations of physical systems from low-resolution data, enabling fast and accurate predictions in engineering applications. However, existing deep-learning based SR methods, require large labeled datasets and lack reliable uncertainty quantification (UQ), limiting their applicability in real-world scenarios. To overcome these challenges, we propose a probabilistic SR framework that leverages the Statistical Finite Element Method and energy-based generative modeling. Our method enables efficient high-resolution predictions with inherent UQ, while eliminating the need for extensive labeled datasets. The method is validated on a 2D Poisson example and compared with bicubic interpolation upscaling. Results demonstrate a computational speed-up over high-resolution numerical solvers while providing reliable uncertainty estimates.
\end{abstract}

\section{Introduction}
A long standing challenge in the engineering sciences is accurately modelling physical systems, most notably when these are described by partial differential equations (PDEs). High-resolution simulations are critical in fields such as automotive and structural engineering \cite{qin2024,hoffmann2005}, where precise modelling of subtle physical behaviours is essential to inform engineering decisions. However, repeated evaluations of high-fidelity simulations using traditional numerical solvers, such as the Finite Element Method (FEM), has high computational costs and significant time requirements. This limitation poses challenges in applications like optimal design, where iterative simulations across varied parameter sets are necessary to achieve optimal configurations, making the process both slow and resource-intensive \cite{wu2021}. With the growing reliance on simulation-based predictions, ensuring computational efficiency alongside accuracy in high-fidelity simulations is paramount. 

To address some of these challenges, researchers have proposed using super-resolution (SR) techniques, from the field of computer vision \citep{dong2015srcnn,shi2016real}, to learn a mapping from low-resolution (LR) images to high-resolution (HR) images. The aim is to obtain high-fidelity solutions of physical fields defined by PDEs by first performing simulations on a coarse grid and then upscaling these coarse predictions to a finer grid.
%
%
Most existing SR techniques for physical simulations focus on learning the mapping using upscaling deep learning models, such as Convolutional Neural Networks (CNNs) \cite{wang2020physicsinformed,gao2021superresolution} and Generative Adversarial Networks (GANs) \cite{bode2021PIESRGAN}, augmented with physics-based constraints.
However, current methodologies exhibit several critical limitations. Firstly, supervised deep learning models such as CNNs and GANs require extensive labelled HR datasets for effective training, the generation of which is computationally expensive. Secondly, existing methods also overlook \textit{predictive uncertainty}, an essential aspect in engineering decision-making. Without accurate uncertainty quantification (UQ) in the HR predictions, the reliability of data-driven decision-making is compromised, especially in engineering systems where uncertainty directly impacts model accuracy and trustworthiness.

In light of this, we propose a Bayesian probabilistic framework for SR of physical simulations, which we call ProbSR. By utilising the statistical Finite Element Method (statFEM) \cite{girolami2021statfem,duffin2021statistical}, coupled with principles from energy-based generative modelling \cite{pang2020}, a conditional distribution of HR predictions given corresponding LR data is obtained. This methodology obviates the need for the training of a complicated upscaling neural network, thereby eliminating the need for large labelled HR datasets. Instead we directly model the residual between HR and LR fields with simpler network architectures. Moreover, the probabilistic nature of the proposed framework inherently facilitates UQ, enhancing the interpretability and reliability of SR predictions. By bypassing the computationally expensive process of simulating full HR solutions, our approach achieves efficient SR while maintaining the robustness necessary for engineering applications. The rest of this paper is organised as follows: in Section~\ref{sec:background} we go over the statFEM methodology as well as Langevin sampling; in Section~\ref{sec:method} we introduce the proposed methodology by dicsussing the model structure, the downscaling network, and the training/inference procedures; in Section~\ref{sec:numerics} we show numerical results demonstrating the effectiveness of the proposed method.

\section{Theoretical Background}\label{sec:background}

In this section, we introduce the relevant background, namely statFEM and Langevin dynamics for MCMC sampling of un-normalised probability distributions. 

\subsection{Statistical Finite Element Method (statFEM)}

The Statistical Finite Element Method (statFEM) is an extension of classical FEM that models physical fields probabilistically. Unlike traditional FEM, which provides deterministic solutions based on discretised governing equations, statFEM focuses on incorporating observational data from engineering systems into finite element models through statistical reasoning by means of Bayesian updating.
In our proposed SR method, we make use of statFEM to provide a probabilistic description of the HR field in the form of a prior distribution.

Consider a standard second-order elliptic PDE
\begin{equation}
    \begin{split}
    \label{eq:pdeexample}
    -\nabla^2 u(x) &=f(x) \ \ \ \ \ x \in  \ \Omega,  \\
    u(x) &= 0 \ \ \ \ \ \ \ \ \ x \in \ \partial \Omega, 
    \end{split}
\end{equation}
where $u(x)$ represents the solution field, $\Omega$ and $\partial\Omega$ denote the PDE domain and boundary, and $f(x)$ is a source term. Following the statFEM approach, we replace the right-hand side term $f(x)$ with a Gaussian Process (GP) prior to represent model misspecification, such that $\xi (x)\sim \mathcal{GP}(f(x),k(x,x'))$, where the kernel function $k(x,x')$ models the spatial correlations of the misspecification. This yields the stochastic PDE%
\begin{equation}
    \label{eq:sdestatfem}
    -\nabla^2 u(x) =\xi(x).
\end{equation}
We discretise \eqref{eq:sdestatfem} with a FEM basis function expansion, $\hat{u}(x) = \sum_{i=1}^{M} u_i \phi_i(x)$, along with test functions,  to obtain the prior distribution,
\begin{equation}
    \label{eq:statfemprior}
    p(\mathbf{u})=\mathcal{N}(\mathbf{u};\textbf{A}^{-1}\textbf{b},\textbf{A}^{-1}\textbf{G}\textbf{A}^{-T}),
\end{equation}
where $\textbf{u}=(u_1,...,u_M)^T$, $\textbf{A}_{ij}=\left \langle \nabla \phi_i,\nabla \phi_j \right \rangle $, $\textbf{b}_j=\left \langle f,\phi_j \right \rangle $ and $\textbf{G}_{ij}=\left \langle \phi_i,\left \langle k(\cdot ,\cdot ),\phi_j \right \rangle  \right \rangle $. GP introduces uncertainty into the PDE solution and is itself characterised by kernel parameters.
Our proposed SR method employs this derived prior as the HR field's prior distribution.

\subsection{Langevin Dynamics}

Langevin dynamics \cite{durmus2019} has emerged as a popular sampling method, which combines elements of both deterministic and stochastic processes to efficiently explore complex probability distributions. This method is based on the Langevin Stochastic Differential Equation (SDE), and holds significant interest across various scientific fields. The Langevin SDE is expressed as
\begin{equation}
  \label{eq:langevinsde}
  dX_t=-\nabla U(X_t)dt+\sqrt{2} dB_t,
\end{equation}
where $U(\cdot)$ represents a potential function and $(B_t)_{t\ge 0}$ is Brownian motion. When using Langevin SDEs to sample from distribution $\pi(x)$, one can set
\begin{equation}
    U(x)=-\log \pi(x).
\end{equation}
Since this is a continuous-time SDE and cannot be directly simulated, discretisation schemes are necessary. Applying a first-order Euler-Maruyama discretization \citep{kloeden1992numerical}, we obtain
\begin{equation}
  \label{eq:langevineuler}
  X_{k+1}=X_k+\gamma\nabla \log \pi(X_k)+\sqrt{2\gamma}W_{k+1},
\end{equation}
where $(W_k)_{k\ge0}$ are drawn from the standard normal distribution and $\gamma$ is a user-selected step size.
Our proposed method uses Langevin dynamics as a sampling scheme to obtain draws from the distribution of HR predictions, thereby performing UQ.

\section{Proposed Methodology}\label{sec:method}
In this section we focus on our proposed methodology. We firstly discuss the probabilistic model structure. We then introduce the downscaling neural network which appears in the probabilistic learning formulation. Finally, we show how our overall SR framework is trained and used for inference.

\subsection{Model Structure}
We will derive the proposed methodology on the Poisson equation, but the method is readily extendable to a wide range of differential equations. We consider again     \eqref{eq:pdeexample}, with varied forcing terms $f_{\theta}(\mathbf{x})$ parameterised by $\theta$.
%
%
We assume a prior distribution for the HR predictions following the structure of the statFEM prior \eqref{eq:statfemprior}
\begin{equation}
    \label{HRprior}
  p(\mathbf{u}^{h} | \theta) = \mathcal{N}(\mathbf{u}^{h}; \mathbf{A}^{-1} \mathbf{b}_{\theta}, \mathbf{A}^{-1} \mathbf{G} \mathbf{A}^{-\top}),
\end{equation}
where $\mathbf{u}^h \in \mathbb{R}^{N_h}$, which is the vector of FEM coefficients for the HR solution field. In the proposed method, we assume $\mathbf{G} = \sigma^2\mathbf{I}$. In numerical solvers, the HR solution is computed by solving $\mathbf{A}\mathbf{u}^{h}= \mathbf{b}_{\theta}$, involving the computationally expensive solving of a system of equations with cost scaling rapidly with the resolution. However, our proposed method circumvents the need for directly solving the linear equation, thereby enabling a more scalable computation of HR solutions.
Next, we define the probability of the LR data given the latent HR solution through a downscaling network $\mathcal{H}_\phi:\mathbb{R}^{N_h}\rightarrow\mathbb{R}^{N_l}$
as
\begin{equation}
   \label{downscaling}
  p(\mathbf{u}^{l} | \mathbf{u}^h, \phi) = \mathcal{N}(\mathbf{u}^{l}; \mathcal{H}_\phi(\mathbf{u}^h), \epsilon^2 \mathbf{I}),
\end{equation}
where $\mathbf{u}^l \in \mathbb{R}^{N_l}$. The training dataset is comprised of only LR data generated from the given equation using various forcing terms $f_{\theta}(x)$. We select $\phi$ by maximizing the marginal likelihood of LR data,
\begin{equation}
  \label{eq:marginallikelihood}
  p(\mathbf{u}^l | \phi, \theta) = \int p(\mathbf{u}^l | \mathbf{u}^h, \phi) p(\mathbf{u}^h | \theta) \, \mathrm{d} \mathbf{u}^h.
\end{equation}
Using Fisher's identity, the gradient of the log-marginal likelihood is given by
\begin{equation}
  \label{eq:gradient}
  \nabla_\phi \log p(\mathbf{u}^l | \phi, \theta) = \int
  \left[ \nabla_\phi \log p(\mathbf{u}^l | \mathbf{u}^h, \phi) \right]
  p(\mathbf{u}^h | \mathbf{u}^l, \phi, \theta) \, \mathrm{d} \mathbf{u}^h.
\end{equation}
Approximating this integral using Monte Carlo sampling, we have
\begin{equation}
  \label{eq:MCgradient}
  \nabla_\phi \log p(\mathbf{u}^l | \phi, \theta) \approx \frac{1}{M} \sum_{m = 1}^M \nabla_\phi \log p(\mathbf{u}^l | \mathbf{u}^{h, (m)}, \phi)  \quad \mathbf{u}^{h, (m)} \sim p(\mathbf{u}^h | \mathbf{u}^l, \phi, \theta).
\end{equation}
Since directly sampling from $p(\mathbf{u}^h | \mathbf{u}^l, \phi, \theta)$ is intractable, we use Langevin sampling~\eqref{eq:langevineuler} to obtain posterior samples as
\begin{equation}
    \label{eq:Langevin}
    \mathbf{u}^h_{k+1}=\mathbf{u}^h_k+\gamma \nabla  _{\mathbf{u}^h} \log p(\mathbf{u}^h_k | \mathbf{u}^l, \phi, \theta)+\sqrt{2\gamma}W_{k+1},
\end{equation}
and selecting $m=1, \hdots,M$ samples from this chain.
We can expand the log gradient term as
\begin{equation}
    \label{eq:logposterior}
    \nabla  _{\mathbf{u}^h} \log p(\mathbf{u}^h | \mathbf{u}^l, \phi, \theta) = \nabla  _{\mathbf{u}^h}   \log  p(\mathbf{u}^l | \mathbf{u}^h, \phi, \theta) + \nabla  _{\mathbf{u}^h}  \log   p(\mathbf{u}^h | \theta).
\end{equation}
The first term in \eqref{eq:logposterior} is the gradient of the likelihood, which can be computed using automatic differentiation (autograd) based on \eqref{downscaling}. The second term is the gradient of the prior, given in \eqref{HRprior}. After rearranging the equation, we avoid the inversion of matrix $\mathbf{A}$ to speed up the procedure,
\begin{align}
    \label{priorarrange}
    \log  p(\mathbf{u}^h|\theta) &\propto
    -\frac{1}{2}(\mathbf{A}\mathbf{u}^h-\mathbf{b}_{\theta})^T\mathbf{G}^{-1}(\mathbf{A}\mathbf{u}^h-\mathbf{b}_{\theta}).
\end{align} 
Thus, the gradient of log prior is
\begin{equation}
    \label{gradientprior}
    \nabla_{\mathbf{u}^h} \log  p(\mathbf{u}^h|\theta) = -\mathbf{A}^T\mathbf{G}^{-1}(\mathbf{A}\mathbf{u}^h-\mathbf{b}_{\theta}).
\end{equation}
This expression only includes multiplication of the matrix $\mathbf{A}$, and $\mathbf{G}^{-1}$ is diagonal, which is easy to compute.
By combining the gradients of the log-likelihood and log-prior into \eqref{eq:logposterior}, we can perform Langevin sampling according to \eqref{eq:langevineuler} to obtain the necessary posterior samples.

\subsection{Downscaling Network}

The architecture of the downscaling network consists of downsampling CNNs designed to learn the residuals between the ground truth LR data and the bicubic downsampled HR data. This approach enables the network to capture fine-grained details lost during conventional downsampling processes. The network is formulated as follows,
\begin{equation}
    \label{eq:networkstructure}
    \mathbf{u}^l = \mathcal{H_{\phi}}(\mathbf{u}^h) = \textit{Bicubic Interpolation}\footnotemark (\mathbf{u}^h) + \mathcal{F}_{\phi}(\mathbf{u}^h),
\end{equation}
\footnotetext{The bicubic interpolation function used is from package \textit{torch.nn.functional.interpolate \cite{paszke2019pytorch}}.}
where $\mathcal{F}_{\phi}: \mathbb{R}^{N_h}\rightarrow\mathbb{R}^{N_l}$, which is the downscaling CNN composed of multiple convolutional layers with nonlinear activation functions and max-pooling layers, designed to progressively extract hierarchical features from the input HR data while reducing spatial dimensions.

\subsection{Training and Inference Procedure}

\subsubsection{Training}

During the training process, the network parameters of the downscaling model, $\phi$, need to be optimised. We first need to generate some LR trainset by FEM using different forcing terms $f_{\theta}$. The following pseudocode outlines the training procedure,

\IncMargin{1em}
\begin{algorithm}[H] 
\SetKwData{Left}{left}
\SetKwData{This}{this}
\SetKwData{Up}{up} 
\SetKwFunction{Union}{Union}
\SetKwFunction{FindCompress}{FindCompress} \SetKwInOut{Input}{input}
\SetKwInOut{Output}{output}
\LinesNumberedHidden
\caption{PDE Super-resolution Algorithm}
\Input{Epoch number T, learning rate $\eta$, initial downscaling model parameters $\phi_0$, high-resolution prior $p(\mathbf{u}^h | \theta)$, low-resolution data $\{\mathbf{u}^{l,(i)}\}_{i=1}^n$, batch size m, posterior sampling step number K and step size s.}
\Output{Downscaling model parameters $\phi_T$}
\BlankLine 
\For{$t = 0 : T-1$}{ 
    \textbf{Mini-Batch:} Sample low-resolution examples $\{\mathbf{u}^{l,(i)}\}_{i=1}^m$\; 
    \textbf{Posterior Sampling:} For each $\mathbf{u}^{l,(i)}$, sample $\mathbf{u}^{h,(i)}\sim p(\mathbf{u}^h | \mathbf{u}^l, \phi, \theta)$ via Langevin dynamics following \eqref{eq:Langevin}\;            
    \textbf{Updating parameters:} $\phi_{t+1}=\phi_{t}+\eta \frac{1}{m} \sum_{i = 1}^m \nabla_\phi \log p(\mathbf{u}^{l, (i)} | \mathbf{u}^{h, (i)},\phi)$
 } 
\end{algorithm} 
\DecMargin{3em}

\subsubsection{Inference}

During the inference phase, we generate HR solutions given a specific LR input by sampling from the posterior $p(\mathbf{u}^h | \mathbf{u}^l, \phi, \theta)$. Thus, once we have trained the downscaling network $\mathcal{H}_{\phi}$,  the SR task can be performed following \eqref{eq:Langevin} until stationarity is reached. This approach allows us to estimate both the mean and covariance of our predictions, thereby enabling UQ within the framework.

\section{Numerical Experiments}\label{sec:numerics}
We now present the numerical experiments and comparisons of ProbSR in a series of test cases.
\subsection{Problem setup}

Take the Poisson problem, where both Dirichlet and Neumann boundary conditions are present,
\begin{align}
    \label{experimenpoisson}
      - \nabla  ^ 2 u(\mathbf{x}) &= f_{\theta}(\mathbf{x}) \ \quad x,y \in [-3,3], \notag \\
       u(\mathbf{x}) &= d \ \quad \quad \ \  \ y \in \{-3,3\}, \\
       \partial_{\mathbf{x}} (u(\mathbf{x}))&=0  \ \quad \quad \  \ \ x \in \{-3,3\}, \notag
\end{align}
where $f_{\theta}(\mathbf{x}) = a\sin(bx)\cos(cy)+b\cos(ax)\sin(cy)+c e^{a\cos(bx)\sin(cy))} + \frac{ax^3-by^3}{x^2+cy^2+1}$. In this case, $a,b,c$, are in the forcing term, and $d$ is in the Dirichlet boundary condition.
A 4-fold SR task is investigated, from resolution $N_l = l \times l$ and $N_h = 4l \times 4l$, where $l$ is the number of points along the side of the square domain. We use $l=40$ in our experiment. The LR training dataset is generated using various source terms and boundary conditions, with 
$ a  \sim \mathcal{U}(-4,4), b\sim\mathcal{U}(-3,3), c\sim\mathcal{U}(0,3), d\sim\mathcal{U}(-2,2)$. 
A total of 1000 training data is randomly generated, which are split into training and testing sets following an 80/20 ratio. 

\subsection{Results}
\begin{figure}[h!]
\centering
\includegraphics[scale=0.41]{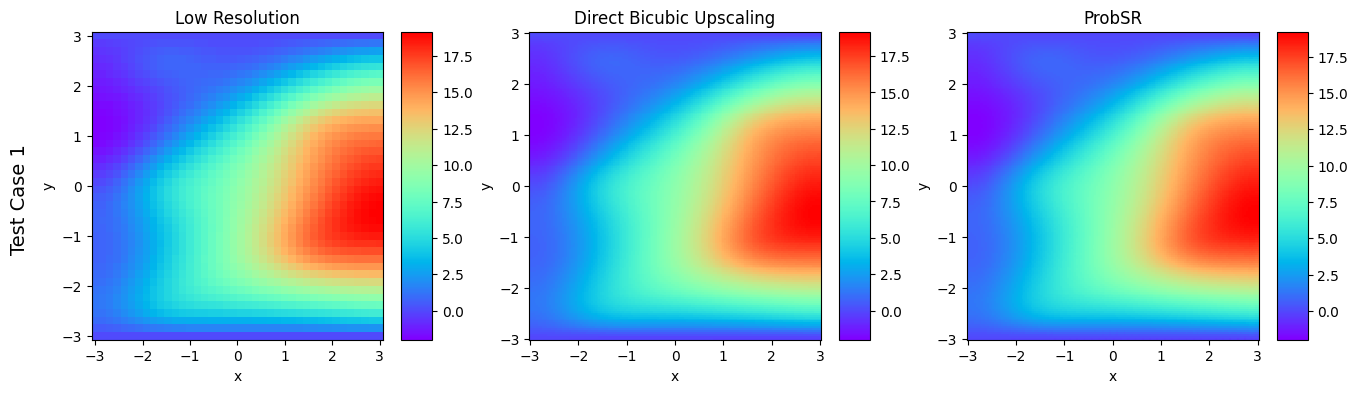}
\includegraphics[scale=0.41]{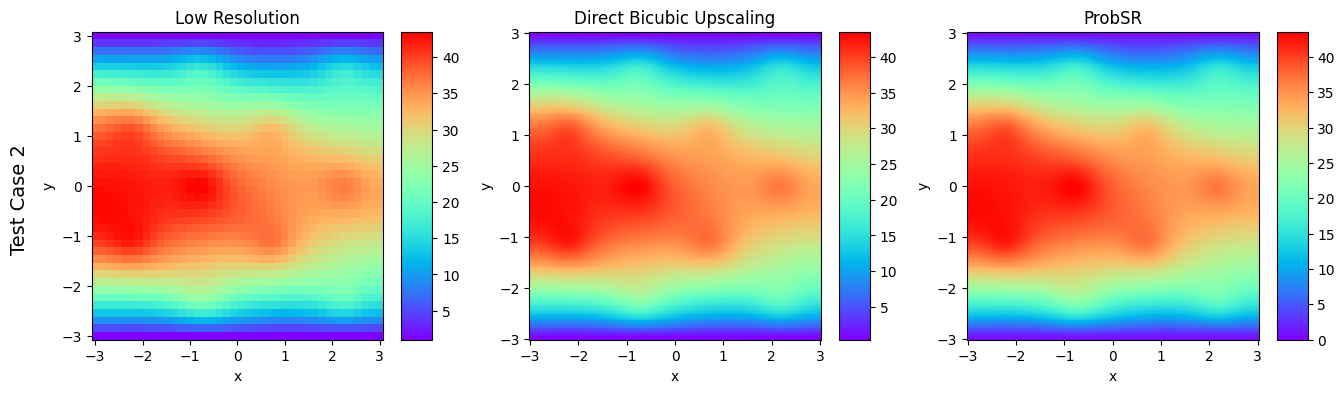}
\caption{The LR solution, and SR results of two test cases using direct bicubic upscaling and ProbSR.}
\label{testcases}
\end{figure}

\begin{figure}
  \begin{minipage}[b]{.45\linewidth}
    \includegraphics[scale=0.46]{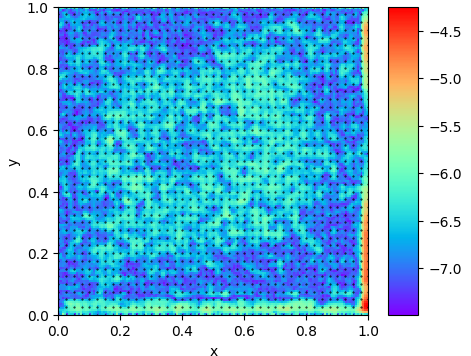}
    \captionof{figure}{Logarithm of the standard deviation field overlaid with black dots representing the original $40 \times 40 $ grid.}%
    \label{UQ}
  \end{minipage}\hfill
  \begin{minipage}[b]{.45\linewidth}
    \centering
    \resizebox{\linewidth}{!}{ 
        \begin{tabular}{c c c}
        \hline
        Test Case & Bicubic Interpolation & ProbSR \\
        \cline{2-3}
        1 & 0.0470 & 0.0412  \\
        2 & 0.4663 & 0.4367  \\
        Mean & 0.1151 & 0.1092  \\
        \hline
        \end{tabular}
    }
    \captionof{table}{MSE of SR results of bicubic interpolation and ProbSR of two test cases and the mean of all test cases.}
    \label{MSE}
\end{minipage}

\end{figure}

 A comparative analysis is conducted between the proposed ProbSR method and the conventional bicubic interpolation. The results for two representative test cases are depicted in Figure \ref{testcases}. In Test Case 1, the parameters are set as $a=b=-2.5, c=1, d=0$, whereas in Test Case 2, the parameters are specified as $a=-3.6, b=2.7,c=2.1,d=1$. The Mean Squared Error (MSE) of both SR methods is reported in Table \ref{MSE}, detailing the errors for the two individual test cases as well as the mean error across all test cases. The results indicate that the proposed method outperforms bicubic interpolation. We attribute this improvement to the incorporation of physical constraints in the prior distribution, which enforces consistency with the governing equation $\mathbf{A} \mathbf{x}=\mathbf{b}$ and improves the accuracy of the reconstructed HR predictions.
%
%
Next, we assess the UQ of the HR predictions for Test Case 1. Specifically, we plot the logarithm of the standard deviation of the Langevin chain in Figure \ref{UQ}, where lower variance is observed at certain locations corresponding to the original LR input points. To further investigate this phenomenon, we overlay the LR grid points ($40 \times 40$) as black dots onto the HR uncertainty field. 
This suggests that in the SR task, the model exhibits higher confidence in its predictions at points mapped from the original resolution, while greater uncertainty arises in the newly interpolated regions.

\begin{figure}[h!]
\centering
\includegraphics[scale=0.45]{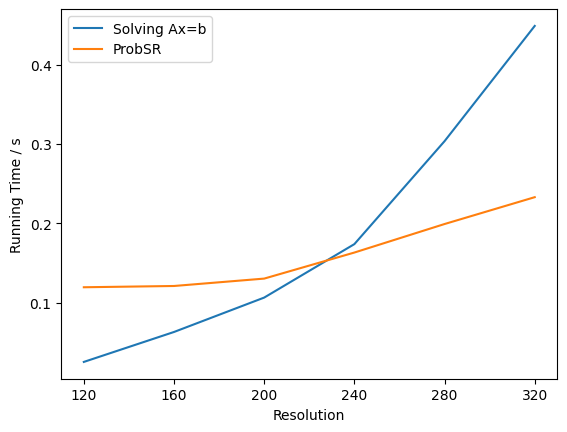}
\caption{Running time of two different methods for generating different resolutions of PDE solution.}
\label{efficiency}
\end{figure}

Finally, the computational efficiencies of the proposed method and traditional numerical methods for solving PDEs are compared. Consider a scenario where a $4l \times 4l$ field is required. We compare the computational cost of the traditional method, which involves solving the linear system $\mathbf{A} \mathbf{u}=\mathbf{b}$, with the ProbSR approach, which first generates a LR solution on an $l \times l$ mesh and subsequently upscales it to a $4l \times 4l$ mesh. The efficiencies at different resolutions, ranging from $120 \times 120$ to $320 \times 320$, are detailed in Table \ref{table:efficiency}. The corresponding plot of running times against required resolutions is shown in Figure \ref{efficiency}. It is evident that traditional solvers perform better at lower resolutions, but as resolution increases, their computational cost grows significantly faster than that of the proposed method. This is due to the need to solve the linear system $\mathbf{Au} = \mathbf{b}$, which scales less favorably than the proposed method as the resolution increases. As a result, the proposed approach is more computationally efficient than traditional numerical methods for high-resolution problems.

\begin{table}[h!]
\centering
\small
\resizebox{\textwidth}{!}{ 
\begin{tabular}{c c c c c c c}
\hline
  \diagbox{Method}{Resolution} & $120 \times 120$ & $160 \times 160$ & $200 \times 200$ &  $240 \times 240$ & $280 \times 280$ & $320 \times 320$ \\ \hline
Numerical solvers & 0.0257 s & 0.0634 s & 0.1067 s & 0.1740 s & 0.3036 s & 0.4487 s \\ 
ProbSR & 0.1197 s & 0.1213 s & 0.1306 s & 0.1634 s & 0.1993 s & 0.2331 s \\ 
\hline
\end{tabular}
}
\caption{Running time of numerical solvers and ProbSR for generating different resolutions of PDE solution}
\label{table:efficiency}
\end{table}

\section{Conclusions and Future Work}

This paper introduces ProbSR, a probablistic framework for enhancing the resolution of physical system simulations. A notable advantage of the proposed method is its independence from HR training data, reducing the computational cost of training. Furthermore, as only a simple downscaling neural network is needed, this approach obviates the need for training a complex upscaling neural network similar to existing literatures, thereby simplifying the overall model architecture.
The experimental results demonstrate that ProbSR outperforms traditional bicubic interpolation by leveraging physics-informed priors, which constrain the HR predictions to better satisfy the underlying governing equations. Furthermore, UQ analysis, conducted via Langevin dynamics sampling, reveals that the model exhibits lower uncertainty at locations corresponding to the original LR grid points, while interpolated regions show higher uncertainty. This finding underscores the reliability of ProbSR in preserving known physical information while acknowledging uncertainty in newly generated HR regions.
Furthermore, the proposed method demonstrates a computational advantage over traditional numerical methods when high-resolution data is required.
This efficiency makes ProbSR a viable alternative for high-fidelity physical simulations, particularly in applications requiring repeated evaluations.
Future work will focus on extending ProbSR to more complex physical systems, incorporating additional domain-specific constraints, and improving the efficiency of the Langevin sampling process by introducing preconditioners.

\section{Acknowledgements}

AGD was supported by Splunk Inc. [G106483] Ph.D scholarship funding. CD and MG were supported by EPSRC grant EP$/$T000414$/$1. AV is supported through the EPSRC ROSEHIPS grant EP$/$W005816$/$1. MG was supported by a Royal Academy of Engineering Research Chair, and Engineering and Physical Sciences Research Council (EPSRC) grants EP$/$T000414$/$1, EP$/$W005816$/$1, EP$/$V056441$/$1, EP$/$V056522$/$1, EP$/$R018413$/$2, EP$/$R034710$/$1, and EP$/$R004889$/$1.

\bibliography{iomac}

\end{document}